%% file: main.tex
\documentclass[letterpaper, 10 pt, conference]{ieeeconf}  % Comment this line out if you need a4paper

\IEEEoverridecommandlockouts                              % This command is only needed if 
\usepackage{amsmath,amssymb,amsfonts}
\usepackage{algorithm, algorithmic}
\usepackage{graphicx}
\usepackage{textcomp}
\usepackage{xcolor}
\usepackage{mathbbol}
\newtheorem{theorem}{Theorem}
\usepackage{booktabs}
\let\labelindent\relax
\usepackage{enumitem}
\usepackage{url}
\newcommand{\mycomment}[1]{}
\begin{document}

\title{Snacks: a fast large-scale kernel SVM solver\\}

\author{Sofiane Tanji$^{1}$ and Andrea Della Vecchia$^{4}$ and François Glineur$^{1,3}$ and Silvia Villa$^{2}$% <-this % stops a space
%\thanks{*This work was not supported by any organization}% <-this % stops a space
\thanks{$^{1}$ICTEAM institute, Université catholique de Louvain, Louvain-La-Neuve, Belgium.
        {\tt\small sofiane.tanji@uclouvain.be}}%
\thanks{$^{2}$MaLGa, DIMA, Universita degli Studi di Genova, Genova, Italy.
        }%
\thanks{$^{3}$CORE, UCLouvain, Louvain-la-Neuve, Belgium.
        }%
\thanks{$^{4}$IIT-Istituto Italiano di Tecnologia, Genova, Italy
        }%
}

\maketitle
\thispagestyle{empty}
\pagestyle{empty}

\begin{abstract}
Kernel methods provide a powerful framework for non parametric learning. They are based on kernel functions and allow learning in a rich functional space while applying linear statistical learning tools, such as Ridge Regression or Support Vector Machines. However, standard kernel methods suffer from a quadratic time and memory complexity in the number of data points and thus have limited applications in large-scale learning. In this paper, we propose Snacks, a new large-scale solver for Kernel Support Vector Machines. Specifically, Snacks relies on a Nyström approximation of the kernel matrix and an accelerated variant of the stochastic subgradient method.
We demonstrate formally through a detailed empirical evaluation, that it competes with other SVM solvers on a variety of benchmark datasets.
\end{abstract}

\section{Introduction}
\input{content/Introduction}

\section{Algorithms and convergence results}
\input{content/Algorithms}

% \section{Convergence results and generalization performance}
% \input{content/Results}

\section{Numerical simulations}
\input{content/Simulations}

\section{Conclusion}
\input{content/Conclusion}

\addtolength{\textheight}{-12cm}   % This command serves to balance the column lengths

\section*{Acknowledgment}
Sofiane Tanji started this work as an intern at MaLGa in Genova, Italy and benefited there from fruitful discussions with Lorenzo Rosasco and Giacomo Meanti. This work has been supported by the ITN-ETN project TraDE-OPT funded by the European Union’s Horizon 2020 research and innovation programme under the Marie Sklodowska-Curie grant agreement No 861137. Silvia Villa also acknowledges the financial support of the European Research Council (grant SLING 819789) and the AFOSR project  FA8655-22-1-7034.

\end{document}

%% file: content/Introduction.tex
Linear statistical learning models are well-studied, convenient to use, to analyze, and they often lead to computationally efficient algorithms \cite{james2013introduction}. However, they assume a linear relation between inputs and outputs and thus often fail to fit complex datasets properly. Kernel methods propose a tractable non-linear extension to linear statistical learning models \cite{scholkopf2002learning, shawe2004kernel}, based on positive-definite kernels. Intuitively, a positive-definite kernel corresponds to a similarity measure between elements of the input space. Such kernels serve as a non-linear mapping which sends the data to a high-dimensional feature space. Computing linear models in this feature space corresponds to computing a non-linear model in the original space. The non-linear function that is learned depends on the kernel function used. As good choices of the kernel function can lead to linearly separable data in the feature space, this approach has been proven successful in many areas such as bioinformatics, signal processing and visual recognition \cite{yang2001face}.
Kernel methods are based on a rigorous mathematical framework that stems from functional analysis and the theory of reproducing kernel Hilbert spaces \cite{aronszajn1950theory, scholkopf2002learning, scholkopf2001generalized} and statistical results can be derived as an extension of linear statistical learning models \cite{NIPS2017_61b1fb3f, calandriello2018statistical}.

Support Vector Machines (SVMs) are historically the first kernel method used for pattern recognition \cite{steinwart2008support, scholkopf2002learning}.
They have been widely used in many contexts such as image classification, text classification \cite{kim2005dimension} or protein fold recognition \cite{ding2001multi}.
Among kernel methods, SVMs are quite popular and represent the state-of-the-art in many contexts. % add references
Unlike many learning algorithms (especially neural networks), SVMs solve a convex optimization problem. This presents a substantial advantage as the structure of the SVM classifier is entirely data-driven once a suitable kernel is chosen.
However, kernel methods require storing a kernel matrix that grows quadratically with the number of samples, limiting SVM solvers to small-scale or medium-scale problems.
This bottleneck has been previously addressed through various approaches.
Exact solvers consider fast optimization algorithms and often decompose the problem by chunking methods \cite{shalev2011pegasos, chang2011libsvm, bottou2007support, wen2018thundersvm}.
Approximate solvers, on the other hand, rely on random features \cite{NIPS2017_61b1fb3f,NEURIPS2018_464d828b, li2019towards} or Nyström subsampling \cite{williams2000using, yang2012nystrom} to reduce the computational cost by considering random projections of the original space, often without leading to any degradation in terms of learning performance \cite{NIPS2015_03e0704b,della2021regularized}.

\subsection{Overview of Snacks}
This paper describes Snacks, a primal kernel SVM algorithm for classification which can handle large-scale datasets.
Our proposed kernel algorithm relies on
\begin{enumerate}
    \item a Nyström approximation to tackle the quadratic complexity inherent to kernel methods and reduce the computational burden \cite{della2021regularized},
    % \item Approximate Leverage Scores (ALS) subsampling which allows the Nyström approximation to be even smaller than with uniform sampling.
    \item an accelerated stochastic subgradient algorithm (ASSG) presented in \cite{xu2019accelerate} to solve the underlying convex optimization problem
\end{enumerate}
% Our 
Worst-case convergence results (Theorem \ref{thm:rate-L2-SVM} and \ref{thm:rate-L1-SVM}) are complemented with numerical experiments on large datasets showing that our algorithm competes with state-of-the-art kernel SVM solvers in terms of training time and classification error.
\subsection{Related work}
Beyond the references cited above, many kernel SVM solvers have been proposed in recent years. They can be divided into two categories based on which formulation they tackle.
Some solvers, such as Pegasos \cite{shalev2011pegasos} (see also \cite{joachims2006training}), tackle the SVM problem in its primal formulation which corresponds to an unconstrained nonsmooth convex optimization problem.
Others, such as LibSVM \cite{chang2011libsvm}, LibLinear \cite{fan2008liblinear}, liquidSVM \cite{steinwart2017liquidsvm}, or \cite{wen2018thundersvm} rely on the dual formulation which is a constrained quadratic program leading to a sparse unique dual solution.
Among all cited solvers, only ThunderSVM and liquidSVM, which work on the dual problem, can tackle datasets with millions of points in reasonable time, which is due to their carefully engineered parallel computations.
Finally, we point out the fact that other kernel methods (such as kernel ridge regression) have been proposed in recent years and can handle billions of points with sound statistical guarantees \cite{NIPS2017_05546b0e, meanti2020kernel, carratino2021park}.
Our work aims at providing a similar scalable algorithm for the SVM problem. The paper is structured as follows. In section \ref{sec:algorithms}, we state the kernel SVM problem, give an overview of existing algorithms tackling both the primal and dual formulations, and present Snacks, our algorithmic solution to the problem. We also report convergence results under classical assumptions and highlight a special case with faster convergence.
% We also provide statistical guarantees on the generalization performance for Snacks.
Section \ref{sec:sim} is dedicated to detailed numerical experiments on various datasets, showing that Snacks is competitive with state-of-the-art solvers on some medium to large-scale classification tasks.

%% file: content/Algorithms.tex
\label{sec:algorithms}
\subsection{Problem formulation}

In classical supervised learning, possible uncertainties coming from the task and the data are taken into account through a statistical model consisting in a pair of random variables $(X,Y)$ taking values in $\mathcal{X}$ (the input space) and $\mathcal{Y} \subset \mathbb{R}$ (the output space) respectively, with joint distribution $P$.
Given a loss function $\ell: \mathcal{Y} \times \mathcal{Y} \rightarrow \mathbf{R}$ that quantifies the closeness between two elements of $\mathcal{Y}$, we define the target of our learning problem, the so called \textit{Bayes function}, as the predictor $f^*: \mathcal{X} \rightarrow \mathcal{Y}$ with minimum expected risk:
\begin{equation}
	f^* \in \arg\min_{f \in \mathcal{L}_2} L(f) = \mathbf{E}_{(X, Y) \sim P} [\ell(Y, f(X))].
\end{equation}

In practice, we cannot directly compute $f^*$ since $P$ is unknown and we only have access to a sampled dataset $S_n = \{(x_i, y_i)\}_{i = 1, \dots, n}$ of random realizations of $(X_1,Y_1),\dots,(X_n,Y_n)$, $n$ i.i.d. copies of $(X,Y)$. From now on, we consider only binary classification problems i.e. where $\mathcal{Y} = \{-1, 1\}$. ERM principle (Empirical Risk Minimization) estimates $f^*$ using these samples. In particular, its output $\hat{f}_n$ is the empirical risk minimizer defined as 
\begin{equation}
\label{eq:initial}
    \hat{f}_n \in \arg\min_{f \in \mathcal{H}} \frac{1}{n} \sum_{i = 1}^n \ell(y_i, f(x_i)) + \lambda \|f\|_{\mathcal{H}}^2
\end{equation}
where $\mathcal{H}$ is a Hilbert space (with induced norm $\|\cdot\|_{\mathcal{H}}$) and we add a regularization term $\lambda \|f\|_{\mathcal{H}}^2$ with parameter $\lambda$.

When $\mathcal{H}$ is a reproducing kernel Hilbert space (RKHS) associated with a positive definite kernel $K: \mathcal{X} \times \mathcal{X} \rightarrow \mathbf{R}$, the representer theorem \cite{scholkopf2001generalized} states that any solution of \eqref{eq:initial} is such that $\hat{f_n} \in \text{span} \{K_{x_1},\dots,K_{x_n}\}$, where $K_{x_i}=K(x_i,\cdot)$. In other words, $\hat{f}_n$ takes the form
\begin{equation}
	\label{eq:repr_thm}
    \hat{f}_n(x) = \sum_{i = 1}^n \alpha_i K(x_i, x).
\end{equation}
where $K$ is now the kernel matrix $K \in \mathbf{R}^{n\times n}$ defined as ${K}_{i,j} = K({x}_i, {x}_j)$.

When $\ell(y_i, f(x_i)) := \max (0, 1 - y_i f(x_i))$, the representer theorem leads to the primal formulation of the kernel SVM problem:
\begin{equation}
    \label{eq:full-primal}
    \hat{\alpha} \in \arg\min_{\alpha \in \mathbf{R}^n} \frac{1}{n} \sum_{i = 1}^n \ell(y_i, [K\alpha]_i)) + \lambda \alpha^\top K \alpha,
\end{equation}

and its dual formulation writes as:
\begin{equation}    \label{eq:dual}
    \tag{Dual L2-SVM}
    \begin{aligned}
    \hat{\alpha} \in \arg\min_{\alpha \in \mathbf{R}^n} \frac 1 2 \alpha^\top K \alpha - \alpha^\top y \\
    \text{ subject to } 0 \leq y_i \alpha_i, \leq \frac{1}{2n\lambda} \forall i = 1, \dots, n
    \end{aligned}
\end{equation}
where $\hat{\alpha}$ is the same $\hat{\alpha}$ as in \eqref{eq:full-primal}.
To tackle the above problem, whether it is in its primal or dual formulation, one needs to store a matrix $K$ of size $n \times n$ which is cumbersome in a large-scale setting. As a remedy, we consider a Nyström approximation of the above problem.

Roughly speaking, we compute a low-rank approximation of the kernel matrix $K$. By subsampling $\{\tilde{x}_1, \dots, \tilde{x}_m\}$, $m$ samples from $S_n := \{x_1,\dots,x_n\}$.

Clearly, taking $m = n$ leads to the original problem \eqref{eq:initial} and with the same optimal solution as in \eqref{eq:repr_thm}. However, to reduce the memory complexity of the initial problem, we will take $m \ll n$. To select the $m$ points, uniform sampling can be used as well as more refined options such as approximate leverage score sampling \cite{rudi2018fast}.

On the reduced subspace $\text{span} \{K_{\tilde{x}_1},\dots,K_{\tilde{x}_m}\}$, the solution of the ERM problem can be written using the representer theorem as follows:

\begin{equation}
    \tilde{f}_m(x) = \sum_{i = 1}^m b_i K(\tilde{x}_i, x)
\end{equation}

and the SVM problem can now be written as follows:

\begin{equation}
    \label{eq:not-parameterized-sub-primal}
    \hat{b} \in \arg\min_{b \in \mathbf{R}^m} \frac{1}{n} \sum_{i = 1}^n \ell(y_i, [\tilde{K}b]_i) + \lambda \langle b, \tilde{K} b \rangle_m
\end{equation}
with $\tilde{K} \in \mathbf{R}^{m\times m}$ defined as $\tilde{K}_{i,j} = K(\tilde{x}_i, \tilde{x}_j)$.

We now reparameterize the above formulation, defining $w = \tilde{K}^{1/2} b$:

\begin{equation}
    \label{eq:parameterized-sub-primal}
    \tag{Primal Nyström L2-SVM}
    \hat{w} \in \arg\min_{w \in \mathbf{R}^m}\frac 1 n \sum_{i=1}^n \ell(y_i, \langle w, \mathbb{x}_i \rangle) + \lambda \|w\|^2
\end{equation}
where 
\begin{equation*}
\label{eq:kernel-embedding}
    \mathbb{x}_i := ((\tilde{K})^{1/2})^\dagger (K(\tilde{x}_1, x_i), \dots, K(\tilde{x}_m, x_i))^\top.
\end{equation*}

The mapping $\mathcal{K}:\mathcal{X} \rightarrow \mathbb{R}^{n \times m}, x \mapsto \mathbb{x}$ above is known as the kernel data embedding. We make extensive use of this embedding in the rest of the paper, as it allows the use of linear solvers for the cost of storing a low-rank approximation of the kernel matrix and considering it directly as the input data.

The next sections describe ways to solve problems \eqref{eq:dual} and \eqref{eq:parameterized-sub-primal}.

\subsection{Optimization methods for dual SVM}
The most common method used to solve \eqref{eq:dual} is Sequential Minimal Optimization (SMO) \cite{platt1998sequential} which corresponds to blockwise coordinate descent. 
% There are several variations and implementations of this algorithm depending on how the Lagrange multipliers are chosen. Early work \cite{platt1998sequential} used greedy heuristics while the LibSVM library \cite{chang2011libsvm} (the most widely used) uses gradient information to conduct the selection.
We briefly break down the SMO algorithm in the following steps:
\begin{enumerate}
    \item Using second-order heuristics, select two training instances $\alpha_u$ and $\alpha_l$ which do not satisfy the complementary slackness condition (see \cite{platt1998sequential} for precisions on the heuristics used)
    \item Minimize the objective over $\alpha_u$ and $\alpha_l$, keeping all other variables fixed.
\end{enumerate}
This method is used in the popular libraries LibSVM \cite{chang2011libsvm} and the more recent ThunderSVM \cite{wen2018thundersvm}.

ThunderSVM presents a slight variation in its implementation. As it makes use of parallelization, it selects a larger subset of variables and solve multiple subproblems of SMO in a batch.

Recall that we work on a variation of the kernel SVM problem using kernel data embedding.
The resulting problem \eqref{eq:parameterized-sub-primal} has the form of a linear SVM problem (the kernel function is the identity) which leads to cheaper gradient computations and faster heuristics to select the two coordinates at each iteration.
In the next section, we test the kernel data embedding variation with a popular linear SVM solver, LibLinear \cite{hsieh2008dual}, which ensures linear convergence using dual coordinate descent (DCD). LibLinear is directly available in Scikit-Learn \cite{pedregosa2011scikit} as the method of choice to solve linear SVM problems.

\subsection{Primal optimization methods for SVM}
Another, less common approach to solve the kernel SVM problem is to solve its primal formulation \eqref{eq:parameterized-sub-primal}.
The problem is convex and nonsmooth due to the hinge loss with cheap explicitly computable subgradients. However, because of the kernel mapping present within the hinge loss, no closed form exists for the proximal operator of the primal objective.
Thus, a natural way to approach the problem is the (stochastic) subgradient algorithm (SSG).
This approach is mainly used in Pegasos \cite{shalev2011pegasos} with a $\tilde{\mathcal{O}}(\frac{1}{\delta \varepsilon})$ convergence rate (with probability $1 - \delta$).
We compare all algorithms previously cited from the complexity point of view in the table below, where $d$ is the size of the feature space, $n$ the number of points, $\lambda$ the regularization parameter and where the convergence rates given correspond to a bound on the number of iterations required to obtain a solution of accuracy $\varepsilon$. For Pegasos (and later on, for the optimization method within Snacks), we report high probability bounds.
\begin{table}[htbp!]
\centering
\caption{Comparison of existing methods to solve the SVM problem}
\begin{tabular}{@{}lccc@{}}
\toprule
\multicolumn{1}{c}{\textbf{Method}} & \textbf{Primal/Dual} & \textbf{Convergence rate} & \textbf{Iter. cost} \\ \midrule
SMO - LibSVM & Dual & $\tilde{\mathcal{O}}(\frac{1}{\varepsilon})$ & $\tilde{\mathcal{O}}( d n )$ \\
SMO - ThunderSVM & Dual & $\tilde{\mathcal{O}}(\frac{1}{\varepsilon})$ & $\tilde{\mathcal{O}}( d n )$ \\
DCD - LibLinear & Dual & $\tilde{\mathcal{O}}(\log \frac{1}{\varepsilon})$ & $\tilde{\mathcal{O}}( d n )$ \\
SSG - Pegasos & Primal & $\tilde{\mathcal{O}}(\frac{1}{\delta \varepsilon})$ & $\tilde{\mathcal{O}}( n )$\\ \bottomrule
\end{tabular}
\end{table}
\subsection{The Snacks algorithm}
In this section, we present Snacks, our algorithm to solve \eqref{eq:parameterized-sub-primal}. Snacks is based on Accelerated Stochastic SubGradient (ASSG), the method proposed in \cite{xu2019accelerate}.
The optimization process starts from an initial guess $w^0$ and generates a sequence of iterates $\{w^k\}_{k = 0}^\infty$. The algorithm consists of an outer loop and an inner loop. An outer iteration (or stage) updates $w^k$ to $w^{k+1}$. During each outer iteration, a sequence of inner iterations leads to a series of updates $w^{k,1}, \dots, w^{k,T}$ (where $T$ is the number of inner iterations).

To update $w^{k,t}$ to $w^{k,t+1}$, we perform the following projected subgradient step:

\begin{equation}
    \label{eq:pssg-update}
    w^{k,t+1} \xleftarrow{} \Pi_{\mathcal{B}(w^{k-1}, D_k)} [w^{k,t} - \eta_k g(w^{k,t}; \xi^{k,t})]
\end{equation}
where $\mathcal{B}(w^{k-1}, D_k)$ is the ball centered on $w^{k-1}$ of radius $D_k$, $\eta_k$ is the stepsize at stage $k$ and $g(w^{k,t}; \xi^{k,t})$ is a stochastic subgradient oracle (with $\xi$ the associated random variable) of the cost function as defined in \eqref{eq:parameterized-sub-primal}.

In our case, $\xi$ picks a random $i$ by uniform sampling and:
\begin{equation}
    g(w; \xi^{k,t}) = \left\{
     \begin{array}{l@{\thinspace}l}
       \lambda w - y_i \mathbb{x}_i & \text{ if } y_i \langle w, \mathbb{x}_i \rangle < 1 \\
       \lambda w & \text{ else.} \\
     \end{array}
   \right.
\end{equation}% écrire la valeur de g(alpha, xi)

At each stage, we shrink the stepsize $\eta_k$ and the ball radius $D_k$ by a factor $\omega$.
Finally, the algorithm is as follows:

\begin{algorithm}[H]
\caption{The Snacks algorithm}\label{alg:snacks}
\begin{algorithmic}[1]
\renewcommand{\algorithmicrequire}{\textbf{Input:}}
\renewcommand{\algorithmicensure}{\textbf{Output:}}
\REQUIRE~~\\
Data: $\mathbb{x}, y, \lambda$,\\ 
Start: $w^0, D_0, \eta_0$,\\
Algorithm parameters: $K, T, \omega$
\ENSURE $w$
\STATE $w, D, \eta$ $\gets$ $w^0, D_0, \eta_0$
\FOR {$k = 1$ to $K$}
    \STATE $c, \overline{w}$ $\gets$ $w$, $w$
    \FOR {$t = 1$ to $T$}
        \STATE $w$ $\gets$ $w - \eta g(w; \xi)$
        \STATE $\overline{w}$ $\gets$ $\overline{w} + \Pi_{\mathcal{B}(c, D)} [w] / T$
    \ENDFOR
    \STATE $w$ $\gets$ $\overline{w}$
    \STATE $D$ $\gets$ $D / \omega$
    \STATE $\eta$ $\gets$ $\eta / \omega$
\ENDFOR
\RETURN $w$
\end{algorithmic}
\end{algorithm}

To the convergence, we state the following theorems which are direct adaptations of the results in \cite{xu2019accelerate}.

\begin{theorem}
\label{thm:rate-L2-SVM}
    For the \eqref{eq:parameterized-sub-primal} problem, the iteration complexity of the Accelerated Stochastic Subgradient Method (ASSG) \cite{xu2019accelerate} for achieving a $\varepsilon$-optimal solution with probability $1 - \delta$ is:
    $$
    \tilde{\mathcal{O}} \left( \frac{\log \delta^{-1}}{ \varepsilon}\right).
    $$
\end{theorem}
This convergence rate is similar to that of the Pegasos solver in terms of high confidence with better dependence to $\delta$.
Moreover, while Pegasos is specifically tailored for SVM with a squared L2-norm as a regularizer, ASSG works for any kind of SVM model (we only need convexity and access to a subgradient).
With other types of SVMs (with non-strongly convex objective) such as L1-SVM (hinge loss with L1-norm as a regularizer), Snacks can also be used. It relies on the sharpness of the objective function and has a linear rate (see \cite{d_Aspremont_2021} for an introduction on how sharpness can control the performance of first-order methods). We leave the evaluation of those variants for future work and only mention one of the corresponding convergence results.
\begin{theorem}
\label{thm:rate-L1-SVM}
    For the L1-SVM problem, the iteration complexity of ASSG for achieving a $\varepsilon$-optimal solution with probability $1 - \delta$ is:
    $$
    \tilde{\mathcal{O}} \left(\log \frac{1}{\delta \varepsilon} \right).
    $$
\end{theorem}

%% file: content/Simulations.tex
\label{sec:sim}
In this section, we evaluate the Snacks solver against several popular SVM solvers, namely LibLinear, Pegasos and ThunderSVM. We focus on binary classification problems as they are naturally handled by the SVM approach.

\subsection{Datasets covered}
To cover a wide variety of scenarios that could arise in kernel SVM learning, we consider the following set of relatively large machine learning datasets, whose size (number of points) is varying between $10^4-10^{10}$:

\begin{table}[htbp!]
	\caption{Sizes of the original datasets and their corresponding kernel matrix}
\begin{center}
\begin{tabular}{lp{2.5em}p{2.5em}p{2.5em}p{2.5em}p{2.5em}}
	    \toprule
		& ijcnn1 &  a9a & MNIST & rcv1 & SUSY \\
		\toprule
        \# of points $n$ & $5\cdot 10^4$ & $5\cdot 10^4$ & $6\cdot 10^4$ & $7\cdot 10^5$ & $5\cdot 10^6$\\
        \# of features $d$ & $22$ & $123$ & $780$ & $4 \cdot 10^5$ & $18$\\
        \midrule
		Dataset (GiB)  & 0.01& 0.05& 0.37& 263 & 0.72 \\
		Matrix $K$ (GiB) & 20& 20& 28.8& 3900 & $2\cdot 10^5$\\
		\bottomrule
\end{tabular}
\label{tbl:results-svm}
\end{center}
\end{table}

One of the dataset, MNIST, comes from the computer vision field and is originally a multiclass dataset. As it contains 10 classes, we classify digit 7 (notoriously hard to classify compared to the other digits) versus all the others. We only consider datasets for which the full kernel matrix (last row in table) can not fit in the RAM of a recent laptop (16 GiB) if stored in float-64 precision.

\subsection{Algorithms, hyperparameters and evaluation metrics}
We compare four SVM solvers on binary classification problems. We recall below the formulation used by each solver and the related optimization method:
\begin{itemize}
    \item LibLinear solves the dual formulation \eqref{eq:dual} on the Nyström subspace (with embedded data) using a coordinate descent method,
    \item ThunderSVM solves the dual formulation of the full problem \eqref{eq:dual} using a parallel coordinate descent method,
    \item Pegasos solves the primal formulation of the Nyström approximation \eqref{eq:parameterized-sub-primal} using a stochastic subgradient method,
    \item Finally, Snacks solves the primal formulation of the Nyström approximation \eqref{eq:parameterized-sub-primal} using an accelerated stochastic subgradient method.
\end{itemize}
For all methods the kernel matrix is pre-computed, except for ThunderSVM because its implementation allows kernel evaluations to be computed on the fly at a neglectable cost.

For all experiments, we use the Radial Basis Function (RBF) kernel. The kernel's bandwidth $\sigma$ is considered as an hyperparameter along with the regularization parameter $\lambda$ of the SVM formulation.

Values of interest are 1) time elapsed for the pre-computation of the Nyström approximation, 2) time elapsed for the training phase (we do not consider the prediction time), 3) classification error on both training set and test set. The classification error corresponds to the percentage of incorrect predictions. As our binarized version of MNIST is a highly imbalanced datasets, we use for it the F1-score metric as a replacement.
We chose to compare classification errors rather than the objective value of the empirical risk minimization problem as the core motivation is to provide a solver for classification problems.

All experiments were run on a Dell Precision 5820 workstation with 128 GB RAM and a 18-core processor running on Ubuntu 20.04 LTS.
Using this experimental setup, we study empirically the following questions :
\setlist{nolistsep}
\begin{enumerate}[noitemsep]
    \item How does Snacks compare in speed and accuracy with other SVM solvers ?
    \item Is Snacks robust to small hyperparameter variations ?
\end{enumerate}
We believe these questions to be crucial to ensure usefulness in real-world usage.

\subsection{Comparison between solvers}
We first compare the four solvers LibLinear, ThunderSVM (dual solvers), Pegasos and Snacks (primal solvers). We provide the time needed for the precomputation of the Nyström approximation as it is not included in the time column in the solver comparison.

To provide the tables below, we follow this protocol : 
\begin{enumerate}
    \item Using LibSVM on the full problem, we perform a gridsearch to tune $(\lambda, \sigma)$, first on a very large grid $\lambda \in [10^{-9}, 10^{1}]$, $\sigma \in [10^{-2}, 20]$ before refining it. We store the average training classification error over 10 runs and consider it to be the "optimal training error".
    \item We run ThunderSVM using hyperparameters $(\lambda, \sigma)$.
    \item We run LibLinear on the sub-problem with increasing subsampling parameter $m$ until optimal training error is reached. We store the corresponding $m$.
    \item We run Pegasos and Snacks using previously found hyperparameters $(\lambda, \sigma, m)$ until the primal accuracy reaches a fixed threshold $\varepsilon = 10^{-6}$.
\end{enumerate}
We use LibSVM and LibLinear to fix hyperparameters because these solvers possess a natural stopping criterion, as opposed to Pegasos and Snacks. We use LibLinear to find optimal $m$ as it is adapted to linear problems such as \ref{eq:parameterized-sub-primal}.

Train test split is done with 5-fold cross validation. Results (average classification error on test set and running time over 10 independent runs) are reported in the tables below.
\begin{table}[htbp!]
\centering
\caption{a9a, $m = 800$. Kernel matrix precomputed in $2.2s$}
\begin{tabular}{@{}rcc@{}}
\toprule
\multicolumn{1}{c}{\textbf{a9a}} & Time (s)        & C-err (optimal = 15.1 \%) \\ \midrule
LibLinear               & 39.0 s          & 15.8 \%                   \\
ThunderSVM                & 2.97 s          & 15.6 \%                   \\
Pegasos                 & 52.0 s          & 20.0 \%                   \\
Snacks                  & \textbf{1.01 s} & \textbf{15.2 \%}          \\ \bottomrule
\end{tabular}
\end{table}

\begin{table}[htbp!]
\centering
\caption{ijcnn1, $m = 5000$. Kernel matrix precomputed in $12.9s$}
\begin{tabular}{@{}rcc@{}}
\toprule
\multicolumn{1}{c}{\textbf{ijcnn1}} & Time (s)        & C-err (optimal = 1.4 \%) \\ \midrule
LibLinear               & 67.1 s          & 1.8 \%                   \\
ThunderSVM                & 31.2 s        & \textbf{1.6 \%}                   \\
Pegasos                 & 1003.5 s          & 3.0 \%                   \\
Snacks                  & \textbf{1.9 s} & \textbf{1.6 \%}         \\ \bottomrule
\end{tabular}
\end{table}

\begin{table}[htbp!]
\centering
\caption{mnist-bin, $m = 3000$. Kernel matrix precomputed in $31.6s$. Metric is F1-score.}
\begin{tabular}{@{}rcc@{}}
\toprule
\multicolumn{1}{c}{\textbf{mnist-bin}} & Time (s)        & F1-score (optimal = 0.998) \\ \midrule
LibLinear               & 19.9 s         & \textbf{0.995}                   \\
ThunderSVM                & 23.6 s       & \textbf{0.995}                  \\
Pegasos                 & 91.8 s         & 0.982                   \\
Snacks                  & \textbf{14.6 s} & 0.985                   \\ \bottomrule
\end{tabular}
\end{table}

\begin{table}[htbp!]
\centering
\caption{rcv1, $m = 1000$. Kernel matrix precomputed in $41.47s$.}
\begin{tabular}{@{}rcc@{}}
\toprule
\multicolumn{1}{c}{\textbf{rcv1}} & Time (s)        & C-err (optimal = 97.1 \%) \\ \midrule
LibLinear               & 1118 s        & 91.1 \%                  \\
ThunderSVM              & 7779 s      & 96.9 \%                \\
Pegasos                 & 61.3 s        & 93.7\%                  \\
Snacks                  & \textbf{7.1 s}& 95.6 \%                  \\ \bottomrule
\end{tabular}
\end{table}

\begin{table}[htbp!]
\centering
\caption{SUSY, $m = 1000$. Kernel matrix precomputed in $74.1s$. ThunderSVM was stopped after 24 hours of training.}
\begin{tabular}{@{}rcc@{}}
\toprule
\multicolumn{1}{c}{\textbf{SUSY}} & Time (s)        & C-err (optimal = 19.8 \%) \\ \midrule
LibLinear               & 9537 s          & 20.2 \%                   \\
ThunderSVM                & NaN        & NaN                   \\
Pegasos                 & 61.2 s          & 21.2 \%                   \\
Snacks                  & \textbf{1.4 s} & \textbf{20.0 \%}          \\ \bottomrule
\end{tabular}
\end{table}

%\begin{table}[htbp!]
%\centering
%\caption{HIGGS, $m = 1000$. Kernel matrix precomputed in $74.1s$. ThunderSVM was stopped after 24 hours of training.}
%\begin{tabular}{@{}rcc@{}}
%\toprule
%\multicolumn{1}{c}{\textbf{SUSY}} & Time (s)        & AUC (optimal = 14.5 \%) \\ \midrule
%LibLinear               & 9537 s          & 20.2 \%                   \\
%ThunderSVM                & NaN        & NaN                   \\
%Pegasos                 & 61.2 s          & 21.2 \%                   \\
%Snacks                  & \textbf{1.4 s} & \textbf{20.0 \%}          \\ \bottomrule
%\end{tabular}
%\end{table}

\noindent Snacks displays a significant speed-up. This was expected when compared to Pegasos as we use an accelerated variant of its optimization method.
The main bottleneck of Snacks lies in its memory requirements: the kernel embedding must be stored fully in memory. This prevented us from tackling even larger scale datasets. When considering the pre-computation time, we obtain total training times similar to ThunderSVM, but only on small datasets (here: a9a, ijcnn1 and mnist-bin). Indeed, ThunderSVM scales badly with size (Snacks has x160 speedup on rcv1 and at least a $\times$1000 speedup on SUSY). This suggests that when trying to achieve optimal accuracy with a small time budget, Nyström subsampling on large-scale datasets is of high interest compared to solving sub-problems repetitively, even in parallel and with on-the-fly kernel computations, as is the case for ThunderSVM.

\subsection{Additional observations}
\begin{figure}[htbp!]
\centerline{\includegraphics[width=0.46\textwidth]{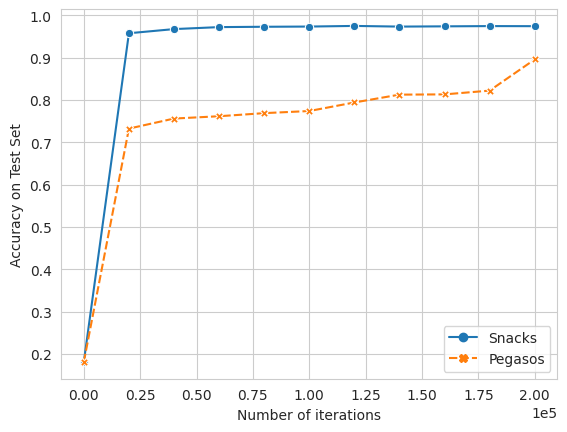}}
\label{fig}
\caption{Accuracy on test set versus number of iterations}
\end{figure}

In Figure \ref{fig}, we run both Snacks and Pegasos for 4 epochs on ijcnn1 dataset and we plot the test accuracy of both classifiers along the number of training iterations.

On this dataset, Snacks converges faster to the optimal value for the same number of iterations. Note that the cost of Pegasos and Snacks' iterations are of similar magnitude (one call to the subgradient oracle and one Euclidean projection) and that the number of iterations for this experiment is significantly larger than during previous experiments (Snacks performs around $40000$ iterations on this dataset).

\subsection{Robustness of Snacks}

Finally, we check the robustness of Snacks against variations of its main hyperparameters.
The heatmap shown here is obtained for the a9a data set. We show how for different choices for the regularization parameter $\lambda$ and for the size $m$ of the random subspace Snacks is relatively robust, as it obtains similar results on a large region around the best values of $m$ and $\lambda$. 

\begin{figure}[htbp!]
\centerline{\includegraphics[width=0.36\textwidth]{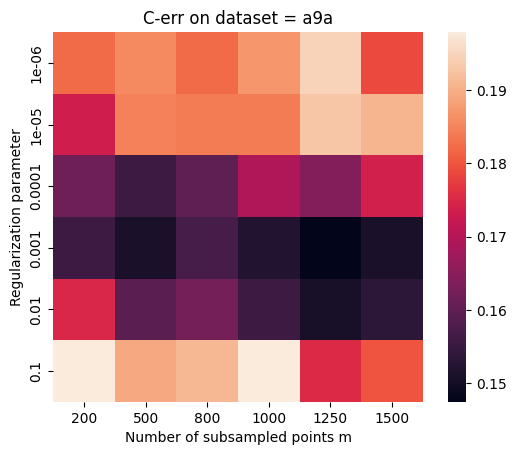}}
\label{fig:acc-vs-m}
\caption{Classification error on test set vs $(m,\lambda)$}
\end{figure}

%% file: content/Conclusion.tex
\label{sec:conclusion}
In this paper, we propose a kernel SVM solver called "Snacks" which performs fast classification on huge scale datasets. The proposed method relies on two key points: reducing the computational burden by solving a Nyström approximation of the original problem and using an accelerated version of the stochastic subgradient method commonly used for kernel SVM solvers.

Numerical experiments support this claim and show a considerable speedup in training time compared to some state-of-the-art SVM solvers. 

Further developments are possible, for example: evaluating Snacks on L1-SVM, providing statistical guarantees (based on \cite{della2021regularized}) for both uniform sampling and Approximate Leverage Scores (ALS) sampling, extending the library for multiclass classification using one-vs-all (OvA) or all-vs-all (AvA) scenarios. We defer these extensions for future work, hoping the Snacks toolbox will benefit practicioners.